\PassOptionsToPackage{table}{xcolor}
\documentclass[]{bytedance_seed}

\usepackage[toc,page,header]{appendix}

\usepackage{minitoc}
\usepackage{amsmath} 
\usepackage{bm} 
\usepackage{booktabs}
\usepackage{graphicx}
\usepackage{makecell}
\usepackage{lipsum}
\usepackage{enumitem}
\usepackage{algorithm}
\usepackage{algpseudocode}
\usepackage{amsmath,amssymb,amsthm}
\usepackage[table]{xcolor}
\usepackage[most]{tcolorbox}
\usepackage{stmaryrd}
\usepackage{todonotes}
\usepackage{amsthm}
\usepackage{parskip} 
\usepackage{wrapfig}
\usepackage{subcaption}
\usepackage{xcolor}
\usepackage{xspace}
\usepackage{marvosym}
\usepackage{amsfonts}       
\usepackage{longtable}      
\usepackage{arydshln}

\newtheorem{definition}{Definition}

\newtheorem{remark}{Remark}
\newtcolorbox{examplebox}[1]{colback=gray!10!white, colframe=black!75, title=#1, fonttitle=\bfseries, arc=0mm, boxsep=1pt, left=2pt, right=2pt, top=2pt, bottom=2pt}
\usepackage{tabularx}
\usepackage{CJKutf8}
\title{DuPO: Enabling Reliable LLM Self-Verification via Dual Preference Optimization}

\author[\heartsuit\spadesuit*]{Shuaijie She}
\author[\spadesuit]{Yu Bao}
\author[\spadesuit]{Yu Lu}
\author[\spadesuit]{Lu Xu}
\author[\spadesuit]{Tao Li}
\author[\spadesuit]{Wenhao Zhu}
\author[\heartsuit(\text{\Letter})]{\authorbreak Shujian Huang}
\author[\spadesuit(\text{\Letter})]{Shanbo Cheng}
\author[\spadesuit]{Lu Lu}
\author[\spadesuit]{Yuxuan Wang}

\affiliation[\spadesuit]{ByteDance Seed}
\affiliation[\heartsuit]{Nanjing University}

\contribution[*]{Work done during an internship at ByteDance Seed}
\contribution[\text{\Letter}]{Corresponding authors}

\abstract{

We present DuPO, a dual learning-based preference optimization framework that generates annotation-free feedback via a generalized duality. 
DuPO addresses two key limitations: Reinforcement Learning with Verifiable Rewards (RLVR)’s reliance on costly labels and applicability restricted to verifiable tasks, and traditional dual learning’s restriction to strictly dual task pairs~(e.g., translation and back-translation).
Specifically, DuPO decomposes a primal task’s input into known and unknown components, then constructs its dual task to reconstruct the unknown part using the primal output and known information~(e.g., reversing math solutions to recover hidden variables), broadening applicability to non-invertible tasks. The quality of this reconstruction serves as a self-supervised reward to optimize the primal task, synergizing with LLMs’ ability to instantiate both tasks via a single model.
Empirically, DuPO achieves substantial gains across diverse tasks: it enhances the average translation quality by 2.13 COMET over 756 directions, boosts the mathematical reasoning accuracy by an average of 6.4 points on three challenge benchmarks, and enhances performance by 9.3 points as an inference-time reranker~(trading computation for accuracy).
These results position DuPO as a scalable, general, and annotation-free paradigm for LLM optimization.

}

\date{\today}
\correspondence{\email{chengshanbo@bytedance.com}(\text{\Letter}), \email{huangsj@nju.edu.cn}(\text{\Letter})}

\begin{document}
\maketitle



\section{Introduction}

Large Language Models (LLMs)~\cite{qwen3,llama3,deepseekv3,GPT-4o,Mistral,gemini25} have shown remarkable progress in tasks like mathematical reasoning~\cite{qwen25math,grpo,AceReason-Math,Big-Math-RL-Verified} and multilingual translation~\cite{seedx,wenhaosurvey,mtpatcher,jiahuanmt}. 
To further enhance their capabilities, reinforcement learning (RL) paradigms like Reinforcement Learning from Human Feedback~(RLHF)~\cite{llama3,deepseekv3,qwen3} and Reinforcement Learning with Verifiable Rewards (RLVR)~\cite{deepseekr1,kimi15,dapo,skywork-or1-2025,orz} have gained traction.
Specifically, RLHF aligns models with human preferences but relies on costly, inconsistent human annotations~\cite{rlaif-compare,human-disagree}. 
RLVR addresses this for objective tasks~(e.g., math, code) via binary rewards from verifiable answers, reducing annotation burdens. 
However, RLVR still depends on external supervision: acquiring verifiable answers remains a bottleneck, limiting scalability. 
Moreover, it struggles with open-ended tasks (e.g., multilingual translation), where single references cannot capture diverse high-quality outputs~\cite{write-zero,bleuproblem}. 
Recent attempts~(e.g., AI-Feedback/RLAIF~\cite{rlaif-compare}, Constitutional AI~\cite{ConstitutionalAI}) merely swap dependencies (human labels $\to$ teacher models/rules), failing to resolve the core bottleneck.

Dual learning~\cite{dualmt} offers a self-supervised alternative by leveraging \textit{task duality} to generate intrinsic feedback:
through paired ``primal'' and ``dual'' tasks (e.g., translation and back-translation~\cite{backtranslation}), models validate outputs via cycle consistency, eliminating reliance on external labels. Given that LLMs possess diverse capabilities from extensive pretraining, they could be trained across various tasks. However, applying this framework to LLMs is non-trivial, which faces two critical challenges:
\begin{enumerate}
    \item \textbf{Limited Duality in Irreversible Tasks}: Most real-world LLM tasks (e.g., creative writing, math reasoning) lack strict invertibility. LLM's output (e.g., a math solution) rarely contains enough information to reconstruct its input (e.g., the original problem), breaking the duality cycle and invalidating self-supervision.  
    \item \textbf{Bidirectional Competence Asymmetry}: LLMs often exhibit uneven performance across primal/dual tasks (e.g., strong at solving math problems but weak at generating problems from solutions). Noisy self-signals from asymmetric tasks hinder optimization, reducing the framework's utility.  
\end{enumerate}
These mismatches render traditional dual learning ill-suited for general LLM optimization, leaving it an open challenge.

In this paper, we propose \textbf{DuPO (Dual Learning-based Preference Optimization)}, a framework that aligns LLM generalization with a (relaxed) duality applicable to general tasks. At its core lies a \textit{generalized duality framework}~(\S\ref{ss:solutions}) built on \textit{complementary relationships}: it decomposes each input $x$ into disjoint known ($x_k$) and unknown ($x_u$) components, then designs the dual objective to reconstruct only $x_u$ from the primal output $y$ and the known input $x_k$, rather than inverting the full input. This framework resolves two asymmetries: it restores sufficient information flow between the primal and dual tasks (task asymmetry) and reduces the complexity burden on the dual task side (capability asymmetry). The formulation naturally synergizes with LLMs: their broad foundational capabilities allow a single model to instantiate both primal and dual tasks without specific architectures, while the dual task converts the model’s outputs into self-supervised reward signals, enabling continual improvement without external annotations. This bidirectional benefit addresses a critical challenge in LLM development: obtaining high-quality feedback for capability enhancement.

We empirically validate DuPO on two diverse and representative tasks: mathematical reasoning and multilingual translation, demonstrating significant and consistent improvements. 
By applying DuPO to one of the strongest translation LLM, Seed-X-7B-Instruct~\cite{seedx}, we demonstrate a significant further performance gain of 2.13 COMET points on the multilingual translation benchmark, bringing the 7B model to performance comparable to ultra large state-of-the-art systems.
In mathematical reasoning, our method yields robust gains across models of varying scales, from 1.5B to 7B parameters; notably, DuPO improves the Qwen3-4B model's score on three challenging mathematical benchmark by 6.4 percentage points. 
Our comprehensive ablation studies confirm that our design, the generalized duality, is crucial for achieving these results. 
Beyond training, DuPO acts as a \textit{reranking mechanism} at inference, boosting performance by 9.3 points without finetuning---enabling smaller models to outperform stronger ultra-large LLM like DeepSeek-R1 even without training.
In summary, DuPO reimagines task duality for non-invertible LLM tasks. It eliminates external annotation reliance, scales across tasks/domains, and enhances both training and inference---offering a scalable path to align LLMs with diverse goals using self-supervised feedback.


\section{Related Work}
\subsection{Preference Optimization for LLMs}
Preference optimization has driven significant advancements in large language models~(LLMs) by aligning outputs with feedback signals, with three dominant paradigms shaping the field:
(1) Reinforcement Learning from Human Feedback~(RLHF)~\cite{instructGPT} has become a cornerstone for aligning LLMs with human preferences.
Its workflow typically involves training a reward model~\cite{skywork-reward-v2,HelpSteer3} on human-annotated preference pairs, then using reinforcement learning (e.g., PPO~\cite{ppo}, GRPO~\cite{grpo}) to optimize the policy model~\cite{qwen3,deepseekv3,llama3}.
While effective for subjective tasks, RLHF faces critical bottlenecks: human annotation is costly to scale~\cite{rlaif-compare}, and consistency across annotators degrades for complex tasks~\cite{human-disagree}, limiting its applicability to large-scale or nuanced scenarios.
(2) Recent work~\cite{mtbench,rlaif-compare,ConstitutionalAI} has leveraged LLM-as-a-Judge to evaluate outputs and provide optimization signals, advancing capabilities in complex tasks. However, the reliability of this paradigm heavily hinges on the judge model's own capabilities and its susceptibility to systematic biases, where evaluations are confounded by various factors such as the presentation order of responses or a preference for certain linguistic styles~\cite{LLMevalNotFair,FalsePromiseLM,ChallengesLLMJudge}.
(3) In response, research has shifted towards exploring Reinforcement Learning from Verifiable Rewards~(RLVR)—a paradigm designed to enhance a model's complex reasoning capabilities in domains like mathematics~\cite{deepseekr1,kimi15,qwen3}. 
By leveraging ground-truth answers as reward signals, RLVR avoids human annotation, but its reliance on verifiable outcomes restricts it to tasks with definitive solutions. 
This makes it ill-suited for open-ended tasks such as multilingual translation, where multiple valid outputs exist and no single ground truth can capture all high-quality responses.

Notably, both paradigms share a fundamental limitation: dependence on external supervision—whether human annotations or pre-defined verifiable answers. This reliance constrains LLMs' adaptability and scalability across diverse tasks, highlighting the need for self-supervised preference optimization mechanisms.

\subsection{Dual Learning}\label{sec:dual-learning}
Dual learning enhances model performance by leveraging intrinsic task symmetry, where a primal task and its complementary dual task mutually provide supervision. 
\citet{dualmt} first introduced dual learning for machine translation, which uses bidirectional tasks (e.g., En$\rightarrow$Zh and Zh$\rightarrow$En) to generate pseudo-labels via back-translation~\cite{backtranslation}, reducing reliance on parallel corpora—a breakthrough for low-resource language pairs.

This framework has since expanded to diverse domains:
\begin{itemize}
    \item \textbf{Cross-modal tasks}: DualGAN~\cite{li2017dualgan} frames image-to-text and text-to-image generation as dual tasks, enforcing cycle consistency to align visual and linguistic representations. 
    \citet{Almost-speechdual} apply a similar principle to text-to-speech~(TTS) and automatic speech recognition~(ASR), enabling joint training with minimal paired data.
    \item \textbf{Knowledge reasoning}: DualTKB~\cite{dualtkb} treats knowledge base path generation and natural language query parsing as symmetric tasks, improving factual consistency via bidirectional validation.
    \item \textbf{Reinforcement learning integration}: \citet{zhang2018deep} designed policy gradient algorithms that transfer rewards between dual tasks, mitigating reward sparsity in low-supervision scenarios.
\end{itemize}

For LLMs, dual learning has enabled capability enhancement. Trans-Zero~\cite{tranzero} uses back-translation to verify semantic preservation in multilingual generation. DualReflect \cite{dualreflect} employs dual tasks (e.g., translation and back-translation) as structured feedback to refine output quality.

However, a critical limitation persists: existing methods require \textit{strict task duality} where primal and dual tasks are mutually invertible (e.g., translation pairs). This restricts application to tasks with ambiguous or non-invertible dual counterparts (e.g., open-ended reasoning, creative writing).
Our work addresses this by reframing dual learning as a preference optimization framework. Instead of relying on explicit task symmetry, we decompose inputs into known/unknown components to construct flexible dual tasks, enabling generalization across diverse tasks without rigid invertibility constraints.
\section{Dual Learning-based Preference Optimization}\label{s:dupo}
In this section, we propose \textbf{Du}al Learning-based \textbf{P}reference \textbf{O}ptimization (\textbf{DuPO}).
Its core objective is to leverage the intrinsic relationships between tasks and their dual counterparts to generate self-supervised rewards, enabling LLMs to improve performance without relying on expensive human annotations or complex handcrafted rules.

\subsection{Task Duality}\label{ss:duality}


We begin by formalizing the task duality between a {primal} task and its {dual} counterpart.
\begin{definition}
\label{def:duality}
Let $\mathcal{X}$ be the input space and $\mathcal{Y}$ the output space.  
A {primal} task is a mapping $\mathcal{T}_p: \mathcal{X} \to \mathcal{Y}$, and a {dual} task is a mapping $\mathcal{T}_d: \mathcal{Y} \to \mathcal{X}$.  
The pair $(\mathcal{T}_p, \mathcal{T}_d)$ is said to form a {dual pair} if they satisfy the {consistency principle}:  
\[
\forall \mathbf{x} \in \mathcal{X},\quad 
d\big(\mathbf{x},\, \mathcal{T}_d(\mathcal{T}_p(\mathbf{x}))\big) \leq \epsilon_{\mathcal{X}},
\]
where $d(\cdot): \mathcal{X} \times \mathcal{X} \to \mathbb{R}^+$ is a domain-specific distance metric,  
and $\epsilon_{\mathcal{X}} \geq 0$ is a tolerance threshold that quantifies acceptable reconstruction errors in each space.
\end{definition}
Leveraging this duality, we can construct a {self-supervised} reward to quantify the quality of a primal-task output.  
Given an input $\mathbf{x} \in \mathcal{X}$ and its corresponding output $\mathbf{y} = \mathcal{T}_p(\mathbf{x})$, we could define reward as
\begin{equation}\label{eqn:dual_reward}
    r(\mathbf{y}) \ \propto\ \exp\!\left(-\lambda \cdot d\ \!\big(\mathbf{x},\, \mathcal{T}_d(\mathbf{y})\big)\right),
\end{equation}
where $\lambda > 0$ controls the sensitivity of the reward to reconstruction error.  
High-quality outputs maximize the expected reward $\mathbb{E}[r(\mathbf{y})]$ by preserving information that is recoverable through the duality cycle. 
This principle has been successfully applied in various domains, including machine translation~\cite{dualmt,tranzero}.

\subsection{Challenges in Dual Learning-based Optimization}\label{ss:challenges}
\begin{figure*}[t]
    \centering
    \subcaptionbox{Non-unique reconstruction of $\mathbf{x}$ from $\mathbf{y}$ \label{fig:challenge-a}}[0.25\textwidth]{
        \includegraphics[width=\linewidth]{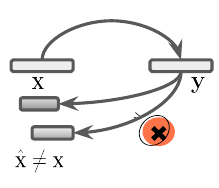}
    }
    \hfill
    \subcaptionbox{Failure to reconstruct $\mathbf{x}$ given $\mathbf{y}$ \label{fig:challenge-b}}[0.25\textwidth]{
        \includegraphics[width=\linewidth]{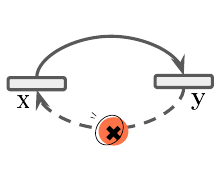}
    }
    \hfill
    \subcaptionbox{Predicting subspaces $\mathbf{B}$ to preserve uniqueness \label{fig:solution-c}}[0.25\textwidth]{
        \includegraphics[width=\linewidth]{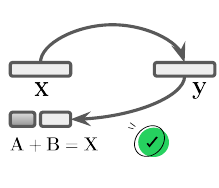}
    }
    \caption{\textbf{Challenges in Dual Learning and Solutions via Relaxed Duality Constraints.} Dilemmas in applying dual learning --- (a) Non-unique reconstruction of $\mathbf{x}$ from $\mathbf{y}$ breaks the closed-loop; (b) Failure to reconstruct $\mathbf{x}$ from $\mathbf{y}$ due to competence asymmetry. Resolutions by relaxing duality restrictions --- (c) Predicting subspaces $\mathbf{B}$ to preserve uniqueness.}
    \label{fig:challenges_and_solutions}
\end{figure*}

While task duality offers a promising self-supervised paradigm, its application to LLM optimization is non-trivial, as it confronts two critical challenges that disrupt the closed-loop information flow. 

\paragraph{\textbf{Challenge I: Limited Duality in Non-Mutually Implicative Tasks.}}
The natural utility of task duality hinges on $\mathcal{T}_p$ and $\mathcal{T}_d$ being mutually implicative --- specifically, the output \(\mathbf{y}\) of $\mathcal{T}_p$ contains sufficient information to reconstruct \(\mathbf{x}\) via $\mathcal{T}_d$, and vice versa. 
This property holds for canonical tasks like machine translation, where \(\mathbf{y}\) (a translation) preserves the semantic content of \(\mathbf{x}\) (the source sentence), enabling $\mathcal{T}_d$ (back-translation) to recover \(\hat{\mathbf{x}} \approx \mathbf{x}\). 

However, most real-world tasks lack this mutual implicativity~(Fig.~\ref{fig:challenge-a}).
Consider mathematical reasoning, where $\mathcal{T}_p$ solves a problem \(\mathbf{x}\) (e.g., ``A box contains 3 red and 5 blue balls; what is the total?'') to produce \(\mathbf{y} = 8\). 
Here, \(\mathbf{y}\) (the total count) is insufficient to uniquely reconstruct \(\mathbf{x}\) via $\mathcal{T}_d$, as \(8\) could answer infinitely many disparate questions, such as ``What is \(10-2\)?'' or ``What is the atomic number of Oxygen?''. 
This underdetermined relationship breaks the duality loop: $\mathcal{T}_d$ cannot reliably recover \(\mathbf{x}\) from \(\mathbf{y}\), making the self-supervised reward (based on \(\hat{\mathbf{x}} \approx \mathbf{x}\)) untrustworthy. 
Such tasks thus require a redefinition of duality beyond direct input-output reversal.

\paragraph{\textbf{Challenge II: Bidirectional Competence Asymmetry.}}
Even for mutually implicative tasks, duality optimization is sensitive to the bidirectional competence of the LLM --- the performance gap between $\mathcal{T}_p$ and $\mathcal{T}_d$. 
If $\mathcal{T}_p$ is strong but $\mathcal{T}_d$ is weak, $\mathcal{T}_d$ may produce noisy \(\hat{\mathbf{x}}\) that distorts the supervision signal~(Fig.~\ref{fig:challenge-b}).  This asymmetry is particularly pronounced in LLMs, where extensive pretraining creates diverse but uneven capabilities across tasks, even within the same domain.

For instance, in machine translation optimization, let \(\mathbf{x} = \) ``The quick brown fox jumps over the lazy dog''~(English) and \(\mathbf{y} = \) ``Der schnelle braune Fuchs springt über den faulen Hund''~(correct German translation). 
A high-quality \(\mathbf{y}\) should enable $\mathcal{T}_d$ to back-translate to \(\hat{\mathbf{x}} \approx \mathbf{x}\). 
However, if $\mathcal{T}_d$ struggles with nuanced vocabulary~(e.g., ``schnelle'' \(\rightarrow\) ``fast'' instead of ``quick''), \(\hat{\mathbf{x}}\) might be ``The fast brown fox jumps over the lazy dog'' --- a divergence from \(\mathbf{x}\) that erroneously penalizes \(\mathbf{y}\) despite its correctness. 



Using separate models for $\mathcal{T}_p$ and $\mathcal{T}_d$, as was common in traditional dual learning, merely sidesteps the challenge of intrinsic competence asymmetry~\cite{wang2022convergencerobustnessadversarialtraining,DualTransL4MT}. This imbalance is still arising from the distinct natures and complexities of the primal-dual tasks, destabilizing the self-supervised feedback loop.

\subsection{Generalized Duality Framework via Complementary Task}\label{ss:solutions}
To address the two-fold challenges of limited duality in non-mutually implicative tasks and bidirectional competence asymmetry, we propose a \textit{generalized duality} that redefines task duality through complementary dependencies. 
It transcends traditional duality’s strict input-output reversal requirement by leveraging \textit{partial and stable dependencies} between task components, enabling robust self-supervised rewarding even for tasks lacking inherent mutual implicativity.

\begin{definition}
\label{def:generalized_duality}
Let the input space \(\mathcal{X}\) of a primal task \(\mathcal{T}_p\) be decomposed into two disjoint subspaces: \(\mathcal{X}_k\)~(known components) and \(\mathcal{X}_u\) (unknown components), such that \(\mathcal{X} = \mathcal{X}_k \cup \mathcal{X}_u\). 
The {primal} task \(\mathcal{T}_p\) is a mapping $\mathcal{T}_p: \mathcal{X} \to \mathcal{Y}$ that maps \(\mathbf{x} \in \mathcal{X}\) to an output space \(\mathbf{y} \in \mathcal{Y}\). 
Its {complementary dual} task \(\mathcal{T}_{cd}\) is a mapping that leverages \(\mathbf{y}\) and the known component \(\mathbf{x}_k\) to reconstruct the unknown component \(\hat{\mathbf{x}}_u \in \mathcal{X}_u\):
\[
\mathcal{T}_{cd}: (\mathbf{y}, \mathbf{x}_k) \mapsto \hat{\mathbf{x}}_u.
\]
The pair $(\mathcal{T}_p, \mathcal{T}_{cd})$ is said to form a {generalized dual pair} if they satisfy the \textit{complementary consistency principle}:
\[
\forall \mathbf{x} \in \mathcal{X},\ \mathbf{y} = \mathcal{T}_p(\mathbf{x}):\ d\big(\mathbf{x}_u, \mathcal{T}_{cd}(\mathbf{y}, \mathbf{x}_k)\big) \leq \epsilon,
\]
where $d(\cdot): \mathcal{X} \times \mathcal{X} \to \mathbb{R}^+$ is a domain-specific distance metric, and \(\epsilon \geq 0\) is a tolerance threshold.
\end{definition}
Leveraging this generalized duality, we can construct a self-supervised reward to quantify the preference of a primal-task output analogously to Def.~\ref{def:duality}. Given an input $\mathbf{x} \in \mathcal{X}$ with decomposition $x=(x_k, x_u)$ and its corresponding output $y=\mathcal{T}_p(x)$, the reward is defined as
\begin{equation}\label{eqn:generalized_dual_reward}
    r(\mathbf{y}) \propto \exp\left(-\lambda \cdot d\left(\mathbf{x}_u, \mathcal{T}_{cd}(\mathbf{y}, \mathbf{x}_k)\right)\right),
\end{equation}
where $\lambda > 0$ controls reward sensitivity. 

Thanks to the generalized duality, we can explicitly use \(\mathbf{x}_k\) (known components) to constrain $\mathcal{T}_{cd}$, enabling stable optimization even when \(\mathbf{y}\) alone is insufficient. 
To highlight how generalized duality resolves the limitations~(\S\ref{ss:challenges}) of classic duality (\S\ref{ss:duality}), we examine a simple two-sum example:

\begin{examplebox}{Example 1: Generalized Duality Feedback for a Two-Sum Task: $A + B$}
\label{ex:math}
The primal task $\mathcal{T}_p: \mathbf{y} \mapsfrom  \mathbf{x}_u + \mathbf{x}_k$ is to compute the sum of two numbers, with its input and output as:\\[-0.3cm]

\quad -- The input $\mathbf{x}$ is decomposed as $\mathbf{x} \mapsfrom (A, B)$, where $\mathbf{x}_k=A$~(a known number) and $\mathbf{x}_u=B$~(an unknown number, without loss of generality).

\quad -- The output $\mathbf{y}$ is the result of sum: $C=A+B$.
\\[-0.3cm]

The complementary dual task $\mathcal{T}_{cd}: \mathbf{x}_u \mapsfrom  \mathbf{y} - \mathbf{x}_k$ is designed to reconstruct the unknown component $\mathbf{x}_u$, using the primal output $\mathbf{y}$~(i.e. $C$) and the known $\mathbf{x}_k$~(i.e. $A$):
\[
  \hat{\mathbf{x}}_u \mapsfrom B' = C - A
\]
Then, we can directly quantifies whether \(B\) (original unknown) and \(B'\) (reconstructed unknown) are consistent as reward signal:  
\[
r(\mathbf{y}) \propto \exp\left(-\lambda \cdot \mathbb{I}(B \neq B')\right).
\]  
Here, \(\mathbb{I}(\cdot)\) is an indicator function: it equals 0 if \(B = B'\) (consistent) and 1 otherwise (inconsistent). This ensures the reward is maximized when \(B\) and \(B'\) match, and reduced otherwise.  
\end{examplebox}

\begin{remark}\label{rem:advantages}
Compared to traditional dual learning, which suffers from strict mutual implicativity (i.e., \(\mathbf{y}\) must fully encode \(\mathbf{x}\)) and bidirectional competence asymmetry, our generalized duality framework offers three fundamental advantages:
\begin{enumerate}
    \item \textbf{Overcomes the Invertibility Constraint.} By redesigning the dual objective from reconstructing the full input \(\mathbf{x}\) to only a selected unknown component \(\mathbf{x}_u\), our framework fundamentally bypasses the stringent requirement of task symmetry. This relaxation is the key to unlocking dual learning for tasks that are inherently non-invertible, where the primal output does not contain sufficient information to recover the entire input.

    \item \textbf{Mitigates the Competence Asymmetry.} The difficulty of the dual task is significantly reduced in two ways. First, the known component \(\mathbf{x}_k\) acts as a strong contextual anchor, constraining the solution space for reconstruction. Second, we can simply yet effectively select an \(\mathbf{x}_u\) that is not only feasibly reconstructible but also act as a faithful reward signal for the primal task’s solution quality~(Appendix~\ref{math-rule}). This directly addresses the ``weak dual'' pitfall and ensures the self-supervised reward is reliable and informative.

    \item \textbf{Enables Broad Applicability.} It unlocks dual learning for a broad class of tasks previously considered unsuitable, including complex domains such as mathematical reasoning, code generation, and dialogue systems where input-output relationships are partial or conditional.
\end{enumerate}
\end{remark}

This generalized duality, therefore, provides a systematic way to overcome the traditional barriers of non-invertibility and competence asymmetry. Our case studies in Appendix~\ref{s:case_study} present concrete examples that illustrate how this process is applied in the multilingual translation and mathematical reasoning scenarios.

\subsection{Preference Optimization}
The core of our Dual Learning-based Preference Optimization (DuPO) framework is to optimize LLMs using duality-derived self-supervised rewards $r(\mathbf{y})$, without external annotations.
The objective is to maximize the expected reward based on the~(complementary) dual task: 
\begin{equation}
    \begin{split}
        \mathcal{J}(\theta) &= \mathbb{E}_{\mathbf{y} \sim \pi_\theta(\mathbf{y} | \mathbf{x})} \left[ r(\mathbf{y}) \right],
    \end{split}
\end{equation}
where $\pi_\theta(\mathbf{y} | \mathbf{x})$ denotes the LLM’s policy (parameterized by $\theta$) for generating output $\mathbf{y}$ given input $\mathbf{x}=(\mathbf{x}_{u}, \mathbf{x}_k)$.
The distance metric $d{(\cdot)}$ design is highly flexible and compatible with various rule-based metrics, enabling application across diverse tasks. 
For example, we could employ BLEU scores for multilingual translation which provide scores from $0$ to $1$, while for mathematical reasoning, we evaluate variable equality, yielding binary rewards.

Notably, DuPO is compatible with various reinforcement learning algorithms~(e.g., PPO~\cite{ppo}, ReMAX~\cite{remax}, REINFORCE{++}~\cite{reinforceplus}), we adopt Group Relative Policy Optimization~(GRPO)~\cite{grpo} in our experiments—for its stability in high-dimensional parameter spaces (critical for LLMs) and compatibility with rule-based rewards.




\section{Experiment}
We validate the efficacy of DuPO on two representative tasks: multilingual translation and mathematical reasoning. 
Below, we detail the experimental setup, datasets, and evaluation metrics for each task, followed by key results.

\subsection{Experiment Setup}
\paragraph{\textbf{Base Model.}}
We evaluate DuPO on a diverse set of strong and popular base models to demonstrate its effectiveness and robustness. For translation tasks, we employ Seed-X-7B-Instruct~\cite{seedx}, one of the strongest open-source translation models. For mathematical reasoning, we select models of varying scales and capabilities, including small-scale yet powerful DeepSeek-R1-Distill-Qwen-1.5B~\citep{deepseekr1} and its larger counterpart DeepSeek-R1-Distill-Qwen-7B, both distilled from the state-of-the-art DeepSeek-R1. We also include Qwen3-4B~\cite{qwen3}, the latest strong small LLM, and the most capable open-source reasoning model, OpenReasoning-Nemotron-7B~\citep{nvreasoner}. These models represent strong and representative baselines within their respective model scales, ensuring comprehensive evaluation. Additionally, we also include some SOTA and impressive ultra-large models like Doubao-1.5/1.6-Thinking~\cite{doubao-think}, Claude-Sonnet4-Thinking, and DeepSeek-R1~\cite{deepseekr1} for comparison.

\paragraph{\textbf{Dataset.}} 
For translation tasks, we focus on 28 languages that are aligned with the language coverage of Seed-X, selecting 1,000 prompts for each language from a multilingual pre-training dataset to create our training prompt set. Additionally, we collect 7,000 parallel data entries across these specified languages to support our experiments from the dev set of Flores-200~\citep{nllb-24}. For mathematical reasoning tasks, we utilize a mixture of publicly available mathematics question datasets\footnote{More details on math data preparation can be found in Appendix~\ref{math-data}.}. These datasets encompass diverse sources and are commonly used for synthesizing supervised fine-tuning data with ultra-large LLMs or conducting reinforcement learning with oracle labels, covering various subjects of competition-level mathematical problems, logic puzzles, and other reasoning tasks.

\textbf{Benchmarks.} To comprehensively evaluate the effectiveness of DuPO, we conduct extensive experiments using the following test sets:
\begin{itemize}
    \item \textbf{Multilingual Translation:} For multilingual translation, we construct our test set by randomly selecting 50 samples for each of the 756 translation directions (among 28 languages) from the testset of Flores\footnote{\url{https://huggingface.co/datasets/openlanguagedata/flores\_plus}}, resulting in a total of 37,800 samples. We will release this dataset for convenient comparison.
    We employ BLEU~\cite{bleu}, COMET~\cite{comet}, and BLEURT~\cite{bleurt} as evaluation metrics. Additionally, we conduct human evaluation on Seed-X-Challenge~\cite{seedx}\footnote{\url{https://github.com/ByteDance-Seed/Seed-X-7B/tree/main/challenge_set}}, a challenging benchmark designed to test the boundaries of LLMs' translation capabilities with diverse linguistic elements across multiple domains. Human experts assess accuracy, fluency, and idiomaticity, scoring translations from Chinese or English to seven languages on a 0-4 scale (higher score denotes better translation quality).
    
    \item \textbf{Mathematical Reasoning:} We evaluate our approach on multiple benchmarks, including AMC23~\cite{amc23}, AIME24~\cite{aime24}, and AIME25~\cite{aime24}, to assess performance in standardized contest environments. For each problem, we sample 32 responses using a temperature of 0.8 and a maximum reasoning budget of 32,000 tokens, then report the average accuracy (Avg@32).
\end{itemize}
Ultra-large models like DeepSeek-R1 and Doubao-1.6-thinking are accessed via their official APIs. More details about training are provided in Appendix~\ref{exp-details}.

\begin{table}[t]
    \begin{minipage}{0.48\textwidth}  
        \centering
        \setlength{\tabcolsep}{2.2pt}  
        \small
        \vspace{-20pt} 
        \begin{tabular}{lcccc}
        \toprule
        \textbf{Model} & \textbf{BLEU} & \textbf{COMET} & \textbf{BLEURT} & \textbf{Avg.} \\
        \midrule
        Qwen3-8B & 21.69 & 84.82 & 65.81 & 57.44 \\
        Doubao-1.5-Thinking & 26.19 & 87.87 & 71.66 & 61.91 \\
        Qwen3-235B-22B & 28.37 & 88.76 & 73.91 & 63.68 \\
        DeepSeek-R1-0528 & 30.21 & 89.16 & 75.03 & 64.80  \\
        \midrule
        
        Seed-X-7B-Instruct & 28.76 & 86.96 & 72.62 & 62.78 \\
        \rowcolor{gray!20} \textbf{w/ DuPO (ours)}  & 30.31 & 89.09 & 74.57 &64.66 \\
        \bottomrule
        \end{tabular}
        \caption{\textbf{Multilingual Translation Performance Across $756$ Translation Directions in $28$ Languages.} DuPO significantly improves all metrics and performs comparably to its strong counterparts (DeepSeek models).
        } 
    \label{tab:translation-metrics} 
    \end{minipage}%
    \hfill  
    \begin{minipage}{0.50\textwidth}  
        \centering
        \vspace{-20pt} 
        \includegraphics[width=\textwidth]{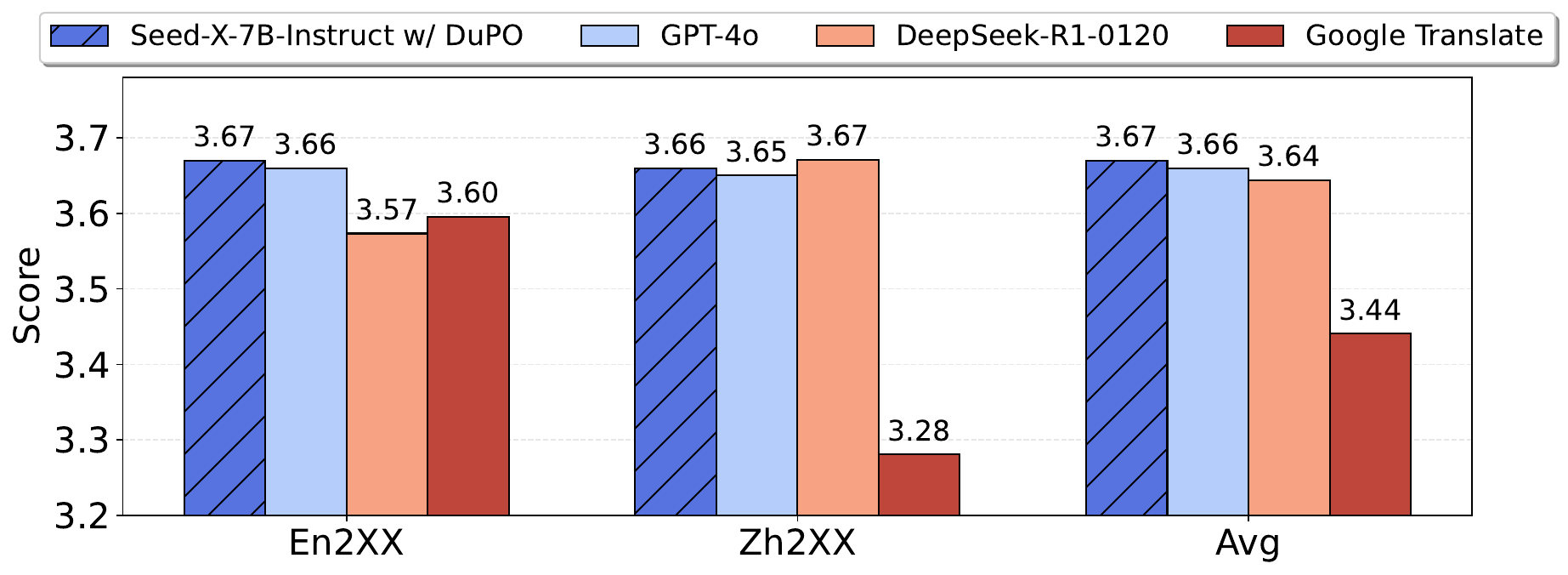} 
        \vspace{-12pt} 
        \captionof{figure}{\textbf{Human Evaluation Scores (0-4) on the Seed-X-Challenge for 14 Language Directions.} DuPO achieves performance comparable to or even surpassing ultra-large models such as GPT-4o and DeepSeek-R1-0120, while significantly outperforming Google Translate.}
    \label{fig:human-mt}
    \end{minipage}
\end{table}




\subsection{Main Results}
\subsubsection{DuPO Boosts LLM's Performance on Various Tasks}
DuPO achieves strong performance on diverse tasks, including multilingual translation and mathematical reasoning. On multilingual translation, DuPO elevates the base model to a state-of-the-art performance level, rivaling and even surpassing significantly ultra large LLM. As detailed in Table~\ref{tab:translation-metrics}, applying DuPO to the Seed-X-7B-Instruct model boosts its performance by 1.55, 2.13, and 1.95 across three automatic evaluation metrics, reaching an average score of 64.66.
This performance even surpasses that of current SOTA closed-source ultra-large language models, such as Doubao1.5-thinking (+2.75) and Qwen3-235B-22B (+0.98), and is on par with the performance of the latest DeepSeek-R1. 
As shown in Figure~\ref{fig:human-mt}, DuPO demonstrates remarkable performance, achieving results comparable to state-of-the-art ultra-large models such as GPT-4o and DeepSeek-R1. Moreover, DuPO substantially outperforms widely-used commercial closed-source systems like Google Translate, showcasing a clear advantage in translation quality as perceived by human evaluators.

\begin{table*}[t]
\centering
\small
\begin{tabular}{lcccccc} 
\toprule 
\textbf{Model} & \textbf{AMC23} & \textbf{AIME24} & \textbf{AIME25}   & \textbf{Average} \\\midrule
DeepSeek-R1-0120 	& 97.7 & 79.8	& 70.0 & 82.5 \\
Claude-Sonnet4-Thinking 	& 97.5 & 82.5 & 70.0 & 83.3  \\
Doubao-1.5-Thinking 	& 99.4 & 86.3	& 73.3 & 86.3 \\
Doubao-1.6-Thinking 	& 98.8 & 88.4	& 83.4 & 90.2  \\
DeepSeek-R1-0528 	& 99.4 & 91.4	& 87.5 & 92.8 \\
\midrule
DeepSeek-R1-Distill-Qwen-1.5B    & 67.5 & 20.0 & 20.0 & 35.8 \\
\rowcolor{gray!20}\textbf{w/ DuPO~(ours)} & 72.5 & 30.0 & 26.7  & 39.7~(+3.9) \\ \midrule
DeepSeek-R1-Distill-Qwen-7B & 85.0 & 56.7 & 36.7 & 59.5 \\
\rowcolor{gray!20}
\textbf{w/ DuPO (ours)} & 90.0 & 63.3 & 40.0 & 64.4~(+4.9) \\ \midrule
Qwen3-4B                     &  95.0 & 70.0 & 66.7 & 77.2  \\
\rowcolor{gray!20}\textbf{w/ DuPO~(ours)} & 97.5  & 83.3 & 70.0 & 83.6~(+6.4) \\\midrule
OpenReasoning-Nemotron-7B      &  95.0 & 83.3 & 73.3 & 83.9  \\
\rowcolor{gray!20}\textbf{w/ DuPO~(ours)} & 97.5  & 83.3 & 90.0 & 90.3~(+6.4) \\ 
\bottomrule
\end{tabular}
\caption{\textbf{Mathematical Reasoning Performances~(\%) on Representative Benchmarks.}  DuPO significantly improves the performances across models with varying base capabilities, enabling Qwen3-4B to outperform DeepSeek-R1-0120 and OpenReasoning-Nemotron-7B to achieve SOTA performance. 
} 
\label{tab:reasoning-metric}
\end{table*}

On mathematical reasoning, the results in Table~\ref{tab:reasoning-metric} clearly demonstrate that DuPO yields consistent and significant performance improvements across all models at different scales and baseline reasoning ability. On the most powerful OpenReasoning-Nemotron-7B model, applying DuPO increased the average score from 83.9\% to 90.3\%, achieving impressive performance.
This trend of significant gains continues on the mid-sized Qwen3-4B model, which saw its average score boosted by 6.4 points from 77.2\% to 83.6\%, even surpassing the ultra-large model DeepSeek-R1-0120. The approach remains remarkably effective on DeepSeek's distilled models as well. Even on DeepSeek-R1-Distilled-Qwen-1.5B, the least reasoning capability among the strong baselines, we still achieved a 3.9-point increase in average accuracy. This directly demonstrates that DuPO is sufficiently robust and stable to enhance the mathematical reasoning capabilities of models consistently. Our framework's robust performance is further validated by concrete examples in multilingual translation and mathematical reasoning (see case studies in Appendix~\ref{s:case_study}).

\subsubsection{DuPO Scales to Various Backbones Effectively}

To validate the robustness and generalization of our proposed DuPO framework, we extend our evaluation to the LlaMA architectural family. Our experiments are conducted on two LlaMA architectural models: LlaMA-3.1-8B~\cite{llama3} and OctoThinker-8B-Hybrid-Base~\cite{OctoThinker}, the latter of which has undergone middle training on mathematical reasoning knowledge. Considering the significant difference of model ability, we select two benchmarks of moderate difficulty, AMC23~\cite{amc23} and MATH500~\cite{math}. For a fair comparison, all models are finetuned using identical training data and settings. 
Results are listed in Table~\ref{tab:dupo-reasoning-others}.

As seen, DuPO's effectiveness is not tied to a specific model architecture; it serves as a robust and generalizable enhancement, delivering significant improvements to diverse backbones regardless of their initial reasoning proficiency. 
DuPO lifts the average score of LlaMA-3.1-8B to 32.1\%, a +24.0 percentage-point gain over the vanilla model, and surpasses SimpleRL-Zoo~\cite{simplerl} (which relies on oracle-labeled answers during training) by 13.1\%. When applied to the OctoThinker-8B-Hybrid-Base~\cite{OctoThinker}, our DuPO approach yields even more impressive performance improvements of +50.0 on AMC23 and +27.4 on MATH500, achieving an average performance of 62.5. 

\begin{table}[t]
    \begin{minipage}{0.45\textwidth}  
        \centering
        \setlength{\tabcolsep}{2pt}  
        \small
        \begin{tabular}{lcccc}
        \toprule
        \textbf{Model} & \textbf{AMC23} & \textbf{MATH500} & \textbf{Average} \\
        \midrule
        LlaMA-3.1-8B                 &  2.5 & 13.6 & 8.1  \\
        \textbf{w/ SimpleRL-Zoo}    &  15.0 & 23.0 & 19.0  \\
        \rowcolor{gray!20}\textbf{w/ DuPO~(ours)} & 20.0 & 44.2 & 32.1 \\ \midrule
        OctoThinker-8B-Hybrid-Base & 5.0  & 42.6 & 23.8  \\ 
        \rowcolor{gray!20}\textbf{w/ DuPO~(ours)} &  55.0 & 70.0 & 62.5  \\ 
        \bottomrule
        \end{tabular}
        \vspace{5pt}
        \caption{\textbf{Performances~(\%) of DuPO on Different Backbone Models.} DuPO even surpasses SimpleRL-Zoo, which utilizes labeled answers as reward. DuPO's potential is further exemplified by OctoThinker, which underwent additional middle training.}
        \label{tab:dupo-reasoning-others} 
    \end{minipage}%
    \hfill  
    \begin{minipage}{0.43\textwidth}  
        \centering
        \includegraphics[width=\textwidth]{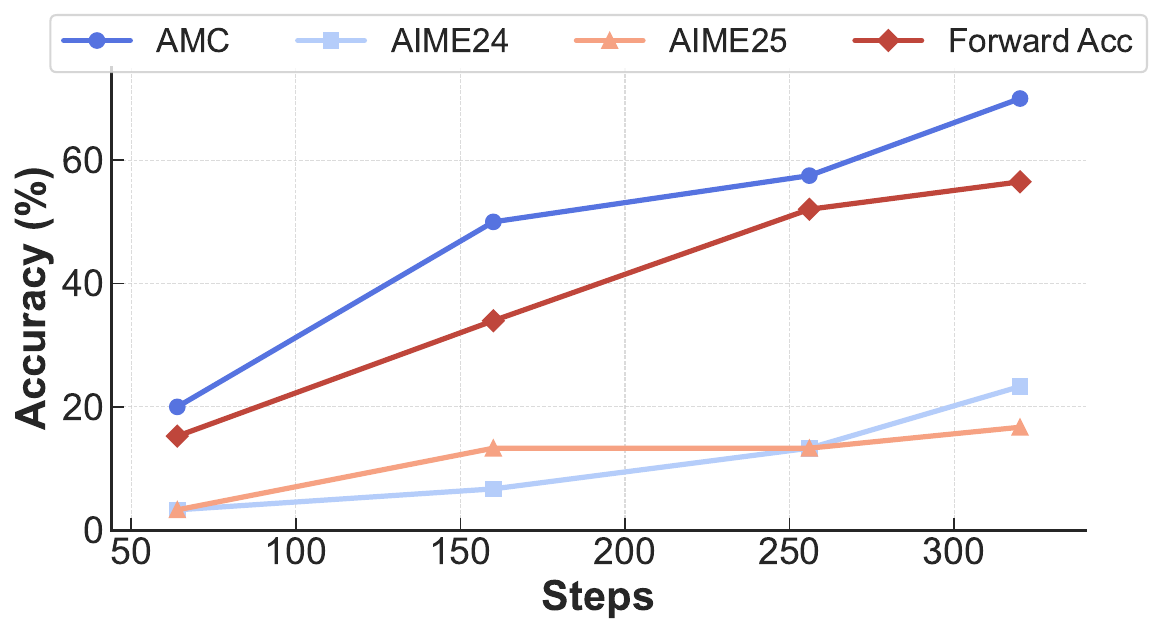} 
        \vspace{-10pt} 
        \captionof{figure}{\textbf{Training Progress of DuPO on Qwen3-4B-Base.} The performance consistently improves on the primal task and the benchmarks.}
        \label{fig:dupo-zero} 
    \end{minipage}
\end{table}


\subsubsection{DuPO Incentivizes Reasoning Capability on Base Model}

We further demonstrate that our DuPO framework can directly elicit and enhance complex reasoning capabilities from a base model. We apply DuPO directly to a base model, without preliminary supervised fine-tuning (SFT) stage activating the reasoing ability. We track the learning dynamics by simultaneously collecting the primal task accuracy (``Forward Acc'') on the training set and its generalization performance on three distinct, unseen challenge test set: AMC23, AIME24, and AIME25.

We can see from Figure~\ref{fig:dupo-zero} that DuPO provides a stable and effective pathway to awaken and generalize the latent reasoning abilities of a base model, validating its utility as a powerful training methodology. 
Specifically, the training dynamics reveal a clear and substantial improvement on the primal task, with the ``Forward Acc'' soaring from a nascent 15.2\% to 56.5\%. 
This upward trajectory provides direct evidence that the reward signal derived from our dual-task serves as an effective guide for enhancing the model's reasoning. More importantly, this acquired skill demonstrates robust generalization. Performance on the unseen test set AMC23 leaped from 20\% to 70\%, with similarly significant gains observed on the AIME24 and AIME25 datasets.

\subsubsection{DuPO Scales Reasoning during Inference without Training}

Beyond serving as a reward signal for RL training, the DuPO mechanism can be naturally applied as a training-free, inference-time reranking strategy to improve the reasoning capabilities of any LLM. The process unfolds in three stages: 1) Similar to the rollout stage during RL process, we could prompt any given policy model to generate diverse reasoning trajectories. 2) For each candidate trajectory, we use its final answer to ask the policy model to solve the corresponding dual question automatic constructed without accessing labeled answer. We could apply more computation by performing $K$~($K=8$ in our experiments) sampling runs on each dual question for a more reliable reward estimate, a practice distinct from RL training. 3) Finally, for each test set question, we select the trajectory with the highest backward accuracy on its dual questions as the final output.

\begin{table*}[t]
\centering
\small
\begin{tabular}{lcccc}
\toprule
\textbf{Model} & \textbf{AIME24} & \textbf{AIME25} & \textbf{Average} \\
\midrule
DeepSeek-R1-0120  & 79.8 & 70.0 & 74.9  \\
Claude-Sonnet4-Thinking & 82.5 & 70.0 & 76.3  \\
\midrule
DeepSeek-R1-Distill-Qwen-1.5B & 20.0 & 20.0 & 20.0   \\
\rowcolor{gray!20} \textbf{w/ DuPO rewarding}  & 53.3 & 24.1 & 38.7~(+18.7)  \\
\midrule
Qwen3-4B & 70.0 & 66.7 & 68.4    \\
\rowcolor{gray!20} \textbf{w/ DuPO rewarding}  & 86.6 & 68.9 & 77.7~(+9.3)  \\
\bottomrule
\end{tabular}
\caption{\textbf{Inference-Time Scaling on Mathematical Reasoning Using DuPO Rewarding~(Backward Acc) for Reranking}. Our method improves the performance of policy models with varying base ability, without requiring additional training.}
\label{tab:scaling} 
\end{table*}

As presented in Table~\ref{tab:scaling}, the experimental results demonstrate that DuPO provides accurate reward signals, effectively guiding models towards correct reasoning, serving as an efficient approach for scaling reasoning capabilities even without training. On the two challenging AIME benchmarks, applying DuPO as a reranking method improves the average performance of Qwen3-4B by 9.3 points, elevating its accuracy from 68.4\% to 77.7\% without any additional training. Notably, the DuPO-enhanced Qwen3-4B surpasses DeepSeek-R1 and Claude-Sonnet4-Thinking~(77.7\% vs. 74.9\%/76.3\% on average). The impact on DeepSeek-R1-Distill-Qwen-1.5B s even more pronounced, with an 18.7 point increase (20.0\% to 38.7\%).

\begin{figure}[t]
    \centering
    \includegraphics[width=0.9\columnwidth]{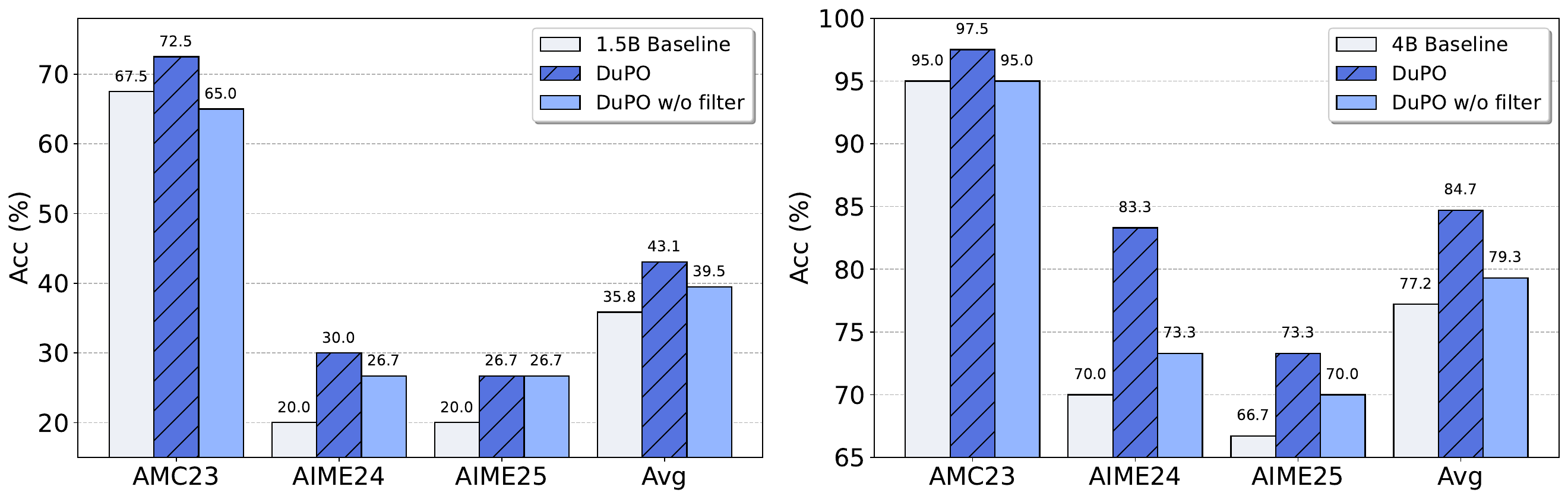}
    \caption{\textbf{Performance Ablation of DeepSeek-R1-Distill-Qwen-1.5B/Qwen3-4B on Mathematical Reasoning}. Our unknown component selection strategy reduces training noise and improves these models' performance across three benchmarks.}
    \label{fig:dumath-ablation}
\end{figure}

\subsection{Effects of Task Duality} 
To thoroughly investigate the effectiveness of our proposed framework and validate how our unknown component selection strategy contributes to achieving better task duality, we conduct an ablation study by maintaining identical experimental settings while removing the unknown component selection mechanism from our dual framework.

As illustrated in Figure~\ref{fig:dumath-ablation}, the results showcase the efficacy of our approach in resolving duality issues. For the 1.5B model, DuPO achieves a remarkable 7.3 percentage point improvement over the baseline. Notably, when we remove data filtering, thereby introducing poorer duality, we observe a significant 3.6 percentage point drop in performance. This pattern not only persists with stronger models. In the case of the 4B model, the benefits of our method become even more pronounced, outperforming the poorer duality variant by an impressive 5.4 points on average. These consistent and substantial improvements across various model sizes provide strong empirical evidence that our component selection strategy is a crucial component of the dual framework, effectively ensuring high-quality task duality and thereby enabling the framework to achieve superior performance.

\section{Conclusion}
We introduce DuPO, a dual learning-based preference optimization framework that eliminates the need for costly human annotations and handcrafted rewards in LLM training.
At its core, DuPO’s innovation lies in a \textit{generalized duality framework} that decomposes and reconstructs input spaces into known and unknown components, addressing critical limitations of traditional dual learning and preference optimization paradigms.
Empirical validation across two diverse, high-stakes tasks confirms DuPO’s versatility and effectiveness.
In mathematical reasoning, DuPO consistently improves performance across model scales from 1.5B to 7B parameters, with notable gains of 6.4\% average accuracy improvement of three benchmarks. 
For multilingual translation, DuPO elevates the 7B-parameter Seed-X model to performance levels comparable to much larger state-of-the-art models, boosting COMET scores by up to 2.13 points across 28 languages and 756 translation directions.
Additionally, DuPO serves as an effective training-free reranking mechanism, enabling smaller models to outperform larger counterparts with up to 9.3 points improvement, bypassing the need for expensive parameter scaling.

DuPO’s model-agnostic design and broad task applicability position it as a scalable solution for annotation-efficient LLM development. 
By harnessing intrinsic task structure to generate self-supervised feedback, it moves beyond the constraints of human supervision and rigid reward engineering—paving the way for more autonomous, adaptable, and cost-effective language model optimization.

\section{Limitations}
Despite the promising results, we acknowledge several limitations of our work that present avenues for future research.
First,  unknown components selection for mathematical reasoning introduces the computational overhead. While this step is crucial for ensuring the quality of the self-supervised reward signal, developing more efficient or even learnable filtering mechanisms could enhance the scalability and practical applicability of DuPO.
Second, our empirical validation is primarily conducted on models of moderate scale. Although DuPO demonstrates consistent improvements across various model sizes, its scalability and effects on significantly larger models remain an open question.
Finally, while we demonstrate DuPO's efficacy on various tasks like mathematical reasoning and multilingual translation, its application to more open-ended and creatively demanding tasks, such as open-ended instruction-following, requires further exploration.








\section*{Acknowledgments}
We extend our sincere gratitude to our colleagues at ByteDance, including Qian Cao\footnote{Qian has already left ByteDance.}, Zhichao Huang, Liyan Kang, Ningxin Peng, Xinghua Qu, Ming Tu, Xiangpeng Wei, Rong Ye, Runsheng Yu, and Zaixiang Zheng, for their valuable advice and insightful discussions, and to Meng Yang and Evaluation Team for their help with the translation evaluation.

\clearpage

\bibliographystyle{plainnat}
\bibliography{main}

\clearpage

\beginappendix

\section{Construction of Dual Question in Math Reasoning}\label{math-rule}
We propose a simple approach for construction of dual question of mathematical reasoning. The algorithm operates on mathematical expressions and performs the following key steps:
\begin{enumerate}
    \item \textbf{Pattern Recognition and Exclusion:} The algorithm first identifies numerical candidates within the expression while excluding numbers in specific contexts: subscripts~($x_1, x_2$), inequality constraints~($x \leq 5$), common exponential bases ($2^n, 10^k$), and function arguments ($f(3)$).
    \item \textbf{Variable Generation and Replacement:} For each valid numerical candidate, the system generates a unique variable identifier of the form $\text{Variable}_{\{str\}}$ where $str$ is a randomly generated lowercase string. The original number is then substituted with this variable.
    \item \textbf{Question Generation of Dual Task:} Using the transformed expression and the original answer, the algorithm constructs inverse problems following templates such as: ``Given that the correct answer is $\{answer\}$, determine the value of $\{variable\}$.''
\end{enumerate}
This methodology enables systematic generation of problem variants while preserving mathematical validity and semantic coherence. 
From a single primal question, multiple dual questions can be derived. To ensure that these dual questions robustly satisfy the properties of duality, we filter the candidates using the following two principles:
\begin{enumerate}
    \item \textbf{Answerability of the Dual Question:} For the set of sampled answers collected for a given primal question, at least one answer must be capable of correctly solving the corresponding dual question.
    \item \textbf{Uniqueness of the Correct Answer:} Among the same set of sampled answers, at most one should correctly answer the dual question.
\end{enumerate}
Taken together, these two principles ensure that for any selected dual question, there is one and only one correct answer within the pool of candidate solutions for the primal task. This establishes the one-to-one correspondence necessary for generating a reliable self-supervised reward signal.

\section{Math RL Dataset Preparation}\label{math-data}
Our dataset preparation process began with the collection of 1,815,942 prompts from various publicly available datasets~\cite{AceReason-Math,Big-Math-RL-Verified,Skywork-OR1-RL-Data,AM-Math-Difficulty-RL}. After deduplication, we obtained 318,649 primal questions and generated 1,059,671 dual questions through our designed steps as discussed above. After that, we employed Qwen2.5-7B-Instruct~\cite{qwen2.5} to sample 32 candidate answers for each primal question and then prompted it to answer the corresponding dual question based on these candidates. Subsequently, we rigorously filtered out all dual questions that failed to meet our predefined principles above. We repeated this sampling and filtering process with Qwen3-4B~\cite{qwen3}, this time with 8 candidate answers per question. The resulting collection of high-quality, diverse mathematical questions along with corresponding dual questions formed our final RL training set, providing a robust foundation for our reinforcement learning tasks in the mathematical domain.

\section{Experiment Details}\label{exp-details}
We presents more details about our training as follows: for the training process, we use a train batch size of 512, mini batch size of 32, sampling temperature of 1.0, and 16 rollouts per prompt, with a learning rate of 1e-6 and gradient clipping set to 1.0. For translation tasks, we set the maximum input length to 2,048 tokens and output length to 4,096 tokens. For mathematical tasks, we use the same input length but extend the maximum output length to 30,000 tokens.


\begin{CJK*}{UTF8}{gbsn} 
\begin{table*}[t] 
    \begin{tabularx}{\textwidth}{p{2.8cm} >{\raggedright\arraybackslash}X}
        \toprule
        \multicolumn{2}{c}{\textbf{Scenario 1: DuPO on Mathematical Reasoning}} \\
        \midrule
        \textbf{Primal Task} & Let $\triangle ABC$ have circumcenter $O$ and incenter $I$ with $\overline{IA}\perp\overline{OI}$, circumradius \textcolor{blue}{\textbf{13}}, and inradius \textcolor{purple}{\textbf{6}}. Find $AB\cdot AC$. (Correct Answer: 468) \\
        \midrule
        \textbf{Dual Task \#1} & Let $\triangle ABC$ have circumcenter $O$ and incenter $I$ with $\overline{{IA}} \perp \overline OI$, circumradius \textcolor{blue}{$\boldsymbol{V_{sk}}$}, and inradius $6$. Find $AB \cdot AC$. Check your work: If the solution for above question is $\boxed{boxed \ answer}$, what must $V_{sk}$ have been? \\
        \midrule
        \textbf{Dual Task \#2} & Let's examine: Let $\triangle ABC$ have circumcenter $O$ and incenter $I$ with $\overline IA \perp \overline OI$, circumradius $13$, and inradius \textcolor{purple}{$\boldsymbol{V_{rj}}$}. Find $AB\cdot AC$. When the solution for above question is $\boxed{boxed \ answer}$, what's the corresponding $V_{rj}$? \\
        \midrule
        \textbf{Candidates} & 
        \begin{tabular}[t]{@{}ll@{}}
            \textcolor{teal}{\textbf{Answer: 468}} & \textcolor{teal}{\textbf{Backward Accuracy: 69.1\%}} \\
            \textcolor{red}{\textbf{Answer: 108}} & \textcolor{red}{\textbf{Backward Accuracy: 0\%}} \\
            \textcolor{red}{\textbf{Answer: 312}} & \textcolor{red}{\textbf{Backward Accuracy: 0\%}} \\
        \end{tabular} \\ 
        \midrule
        \midrule
         \multicolumn{2}{c}{\textbf{Scenario 2: DuPO on Machine Translation~(MT)}} \\
        \midrule
        \textbf{Primal Task} & Translate to Chinese: As knowledge of Greek declined, the West found itself cut off from its Greek philosophical and scientific roots. \\
        \textbf{Reference} & 随着希腊知识的衰落，西方脱离了其希腊哲学和科学根源。 \\
        \midrule
        \textbf{Primal MT \#1} & 随着希腊语知识的衰落，西方发现自己与希腊的哲学和科学根源失去了联系。\textcolor{teal}{\textbf{(BLEU: 45.85)}} \\
        \textbf{Dual MT \#1} &  As knowledge of Greek declined, the West found itself cut off from its philosophical and scientific roots in Greece.\textcolor{teal}{\textbf{(BLEU: 82.07)}} \\
        \midrule
        \textbf{Primal MT \#2} & 随着对希腊语的了解逐渐消失，西方发现自己与希腊哲学和科学根源隔绝开来。\textcolor{red}{\textbf{(BLEU: 28.65)}} \\
       \textbf{Dual MT \#2} &  As understanding of the Greek language gradually fades, the West finds itself cut off from the roots of Greek philosophy and science.\textcolor{red}{\textbf{(BLEU: 16.11)}} \\
       \bottomrule
    \end{tabularx} 

  \caption{\textbf{Case Studies of DuPO on Mathematical Reasoning and Machine Translation.} DuPO validates each candidate's quality through a corresponding dual task, reliably identifies \textcolor{teal}{\textbf{the superior solution}} over \textcolor{red}{\textbf{inferior ones}}.}
  \label{tab:case_study}
\end{table*}

\end{CJK*} 

\section{Case Study}\label{s:case_study}
To illustrate the efficacy of our DuPO approach, we present two representative scenarios in Table~\ref{tab:case_study} that demonstrate how DuPO provides a reliable reward signal across diverse domains.

\textbf{Scenario 1: Mathematical Reasoning Validation.} In mathematical reasoning, DuPO derives dual task questions from the primal task question where key numerical parameters are replaced with variables, and the model tries to work backwards conditioned on candidate answers. When given a geometry problem about triangle properties, three candidate answers are sampled: 468, 108, and 312. DuPO automatically derives two dual questions by replacing the circumradius (13) and inradius (6) with variables, asking the model to deduce these values from the proposed answer. The candidate answer 468 achieves 69.1\% accuracy on dual task, while the incorrect answers (108 and 312) totally fail to answer the dual task.

\textbf{Scenario 2: Machine Translation Quality Assessment.} For translation tasks, DuPO leverages reverse direction translation as the dual task to evaluate translation quality. Given an English sentence about Greek philosophical decline, two Chinese translation candidates are generated and subsequently back-translated to English. The first translation achieves a BLEU score of 45.85 in the forward direction and 82.07 in the back-translation, demonstrating semantic preservation and translation fidelity. In contrast, the second candidate shows degraded performance with BLEU scores of 28.65 and 16.11, respectively, indicating semantic drift and poor translation quality.

These case studies validate DuPO's core hypothesis: high-quality solutions maintain consistent information across dual task formulations, while inferior solutions exhibit significant degradation. This dual validation mechanism provides a robust framework for automatic quality assessment without requiring ground truth labels.

\end{document}